# Generating Multimodal Images with GAN: Integrating Text, Image, and Style

Chaoyi Tan*

Northeastern University, tan.chaoyi@icloud.com

Wenqing Zhang

Washington University in St. Louis, wenqing.zhang@wustl.edu

Zhen Qi

Northeastern University, garyzhen79@gmail.com

Kowei Shih

Tsinghua University,  skw19@tsinghua.org.cn

Xinshi Li

Montclair State University, Xinshili9@gmail.com

Ao Xiang

Northern Arizona University, ax36@nau.edu

In the field of computer vision, multimodal image generation has become a research hotspot, especially the task of integrating text, image, and style. In this study, we propose a multimodal image generation method based on Generative Adversarial Networks (GAN), capable of effectively combining text descriptions, reference images, and style information to generate images that meet multimodal requirements. This method involves the design of a text encoder, an image feature extractor, and a style integration module, ensuring that the generated images maintain high quality in terms of visual content and style consistency. We also introduce multiple loss functions, including adversarial loss, text-image consistency loss, and style matching loss, to optimize the generation process. Experimental results show that our method produces images with high clarity and consistency across multiple public datasets, demonstrating significant performance improvements compared to existing methods. The outcomes of this study provide new insights into multimodal image generation and present broad application prospects.

CCS CONCEPTS

Computer graphics~Image manipulation~Image processing

Computing methodologies~Artificial intelligence ~Computer vision

---

* Place the footnote text for the author (if applicable) here.

## 1 INTRODUCTION

Generative Adversarial Networks (GANs) have revolutionized image generation by using adversarial learning between a generator and discriminator to create realistic images. GANs are widely used in image synthesis, restoration, and style transfer[1]. However, as image generation tasks become more complex, single-modality methods are insufficient for diverse applications. Multimodal generation, which integrates text, images, and style, has gained attention for generating images with specific content and aesthetics. Each modality offers unique information: text provides semantics, images offer visual references, and style defines the overall look. The challenge lies in combining these elements while maintaining consistency. Current methods often lack flexibility in style or semantic guidance. This paper presents a GAN-based multimodal image generation method that combines text encoding, image feature extraction, and style integration. Our method uses adversarial, text-image consistency, and style matching losses to optimize the process, showing superior performance across datasets in terms of semantic alignment, visual quality, and style diversity[2].Yang et al.[3] proposed an enhanced U-Net model for remote sensing image segmentation, integrating the SimAM and CBAM attention mechanisms. Zhu et al.[4] proposed an innovative OOD detection approach enabling robust detection without the need for labeled in-distribution data. Dong et al.[5]proposed the integrates reinforcement learning with graph neural networks. Gong et al.[6] introduced a hybrid CNN-LSTM model with attention mechanisms. Shen et al.[7] proposed a CNN-LSTM-Attention model for temperature forecasting, showcasing the effectiveness of attention mechanisms. Ke et al. [8] designed a hybrid model combining a back-propagation neural network and genetic algorithm.

## 2 GAN-BASED MULTIMODAL IMAGE GENERATION: CONCEPTS AND METHODS

In computer vision, Generative Adversarial Networks (GANs) have emerged as a powerful tool for image generation, particularly for multimodal tasks. Multimodal generation involves combining inputs like text, images, and styles to produce images that align with these inputs[9]. We propose a conditional GAN (cGAN) architecture, where text input guides the image generation process. <Figure 1> illustrates this architecture, which combines text descriptions, image features, and style information to create images that meet user requirements[10].The generator is driven by text descriptions, which are encoded into feature vectors and combined with random noise to generate new images. The text is transformed into a latent feature vector through an encoder, carrying semantic information that guides the generation. Random noise is introduced along with the text features, enabling the generator to produce images that align with the text's semantic requirements[11].

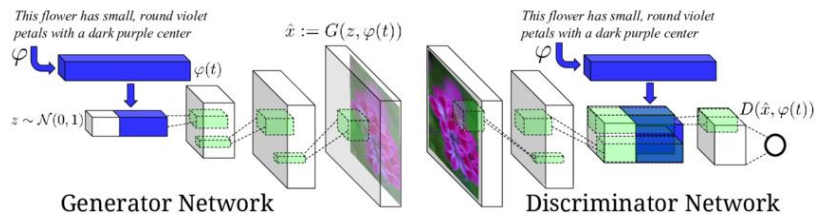

Figure 1: Text-Conditional Multimodal GAN Architecture

The generator processes inputs through multiple convolutional layers, progressively generating high-quality images that match the text description. The convolutional and deconvolutional layers help generate image details, transitioning from low to high resolution. The text feature plays a crucial role, ensuring the image maintains semantic consistency with the original description[12]. The generator also introduces style information, allowing control over the image's overall style, resulting in diverse visual effects.The discriminator evaluates the generated images, determining whether they are real or fake, and assessing consistency with the text description. The discriminator analyzes both the generated image and text features to judge authenticity and alignment with the text, helping the network learn to distinguish between real and generated images. This forces the generator to improve the quality of its output continuously[13].To enhance the discriminator's accuracy, we incorporate multiple loss functions, including adversarial loss, text-image consistency loss, and style matching loss. These ensure the generated images not only appear realistic but also align with both the text and style inputs. Yu Q et al. utilized unsupervised learning models to optimize systems.[14]The adversarial loss improves image realism through adversarial training, while the text-image consistency loss ensures the image matches the input text's semantic content. The style matching loss ensures the generated image reflects the reference style.By integrating these loss functions, the GAN produces realistic, multimodal images, achieving true multimodal integration[15]. Our text-conditional multimodal GAN offers several advantages over traditional image generation methods. It generates images with higher semantic consistency, making it suitable for text-to-image generation and personalized artistic creation[16]. The style integration module allows for diverse stylistic effects, enhancing the utility and diversity of the images. Additionally, multimodal consistency loss ensures coherence across modalities, supporting complex generation tasks.In conclusion, our GAN model integrates text, image, and style inputs to generate high-quality images that meet diverse, multimodal requirements, offering new solutions in computer vision and practical applications like artistic creation and product design[17].

## 3  MODEL DESIGN

### 3.1  GAN-Based Multimodal Image Generation Architecture

This section introduces the proposed multimodal image generation architecture, which combines Generative Adversarial Networks (GAN), diffusion models, and Transformers. To generate high-quality images that match the input text descriptions, we utilize pixel-level and latent space processing with diffusion models, along with the Transformer's ability to integrate multimodal information effectively[18]. <Figure 2> shows the design flow of the whole model and the working principle of different parts.

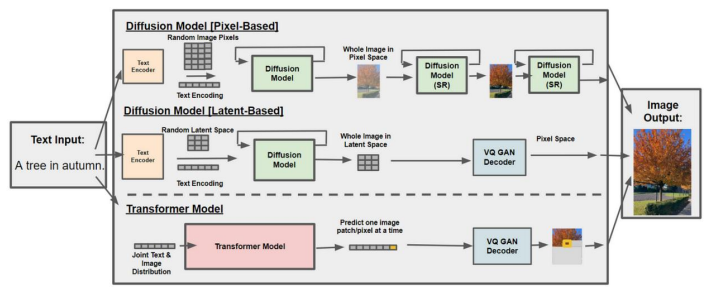

Figure 2:  illustrates the overall model design and the functioning of each component.

In this architecture, text descriptions serve as the primary input, converted into feature vectors by a text encoder. These vectors are used as conditions for generating images, with either diffusion models or Transformers guiding the process to match the input text. The diffusion model, acting as the primary generation component, progressively generates images in both pixel and latent spaces. The pixel-space diffusion model begins with random pixel values, generating the full image through a gradual diffusion process, followed by super-resolution (SR) processing for high-resolution output. The latent-space diffusion model generates images in a low-dimensional latent space, which are then decoded into pixel-space images via a VQ-VAE decoder. This method efficiently handles complex image generation tasks while reducing computational costs.In both models, text features serve as conditional information throughout the generation process, ensuring that the generated images align with the text descriptions. Additionally, the Transformer architecture generates images pixel-by-pixel or block-by-block, capturing long-range dependencies between text and image[19]. Finally, the VQ-GAN decoder converts latent representations or pixelated predictions into high-quality images, maintaining both diversity and visual consistency.This multimodal architecture effectively combines the strengths of diffusion models and Transformers, producing semantically consistent, high-quality images with rich detail[20].

### 3.2  Image Feature Extraction and Style Integration

In multimodal image generation tasks, apart from generating content based on text, image feature extraction and style integration are also critical factors in ensuring that the generated images meet user requirements. The image feature extraction module is responsible for extracting key visual information from input reference images, while the style integration module controls the overall visual style of the generated images. This ensures that the final output images match the text description in content and exhibit diversity and personalization in style.In our architecture, we designed detailed modules for image feature extraction and style integration, ensuring that the generated images maintain high quality and diversity. To enhance the visual consistency of generated images, we use a convolutional neural network (CNN) to extract features from input reference images[21]. Through multiple convolutional operations, the CNN effectively captures local and global features of the image, such as color distribution, shape contours, and spatial structure. These features are then embedded into the latent space of the generator, serving as conditional information during the generation process. This allows the generation network to reference the content features of the input image during generation, ensuring that the generated images retain some similarity to the original image, particularly in terms of visual elements.In the diffusion models for pixel and latent space, the image feature extraction module plays a crucial role. The pixel-space diffusion model relies on precise pixel information, so the feature extraction of the reference image directly affects the details of the generated image[22]. In the latent-space diffusion model, the image features are embedded into the latent space, guiding the generator to learn the overall structure and layout of the image more effectively.The style integration module is designed to enhance the diversity and flexibility of the generated images in multimodal generation tasks. The generated images need to match the content of both the text and reference images, but also need style adjustments to meet user requirements for different visual styles. For this reason, we introduce a style encoder that extracts a feature vector representing the overall visual style by analyzing the input style image. Style features include visual elements such as color, texture, and lighting, which are integrated into various stages of the generation process, influencing the visual effects of the image.Typically, the style integration module is deeply embedded in the generator within the GAN architecture. Specifically, the style feature vectors are combined with text and image features and injected into different layers of the generator to ensure that the generated image reflects style variations at different levels. This allows the generation network to produce images with diverse styles while maintaining content consistency. For example, in different scenarios, users can generate images with different tones, artistic styles, or lighting effects through the style integration module, greatly enhancing the model's utility.To ensure that the generated images match the specified style, we designed a style matching loss function. This loss function quantifies the difference between the generated image and the target style and uses backpropagation to guide the generator to produce images that match a specific style. The design of the style matching loss typically combines perceptual loss and style loss, where perceptual loss ensures that the structural features of the image remain consistent, and style loss focuses on maintaining visual style consistency.Through this design of feature extraction and style integration, our model generates high-quality images that match both text descriptions and various visual styles. This generation method has wide potential applications, such as personalized artistic creation, automated design, and content generation[23].

### 3.3 Loss Functions and Optimization

To achieve high-quality multimodal image generation, designing appropriate loss functions is a critical step. Our proposed Generative Adversarial Network (GAN) architecture incorporates multiple loss functions, including adversarial loss, text-image consistency loss, and style matching loss. These combined loss functions ensure that the generated images exhibit consistency in content, style, and semantics.Adversarial Loss is the core component of GAN. It trains both the generator and the discriminator, allowing the generator to produce realistic images while the discriminator differentiates between real and generated images. For the generator G and discriminator D, the training objective is a minimax game: the generator tries to minimize the discriminator's ability to distinguish real from generated images, while the discriminator maximizes its ability to make this distinction.The adversarial loss function is defined as shown in Formula 1:

$$L_{GAN}(G, D) = E_{x \sim pdata(x)}[\log D(x)] + E_{z \sim pz(z)}[\log(1 - D(G(z, \varphi(t))))] \quad (1)$$

Where x represents real images, z is the random noise vector, and $\varphi(t)$ is the text feature vector generated from the text encoder. The generator G produces the generated image $G(z, \varphi(t))$, and the discriminator D tries to differentiate between real images x and generated images. By optimizing the adversarial loss, the generator gradually improves the realism of its outputs.In multimodal image generation, ensuring the generated images align with the input text description is crucial. For this, we designed the Text-Image Consistency Loss, which quantifies the semantic similarity between the generated images and text descriptions. Using a CNN-based image encoder and a Transformer-based text encoder, we map both image and text to the same feature space and compute their similarity.The text-image consistency loss is defined as shown in Formula 2:

$$L_{txt-img} = \| \phi_I(G(z, \varphi(t))) - \phi_T(t) \|_2^2 \quad (2)$$

Where $\phi_I$ is the image encoder, $\phi_T$ is the text encoder, $G(z, \varphi(t))$ is the generated image, and t is the input text description. This loss minimizes the Euclidean distance between the feature vectors of the generated image and the input text, ensuring semantic alignment between the two.To generate images with specific styles, we designed the Style Matching Loss. This loss ensures that the generated image's style matches the reference style image. It draws inspiration from the perceptual loss and style loss proposed by Gatys et al., comparing high-level features between the generated image and the target style image.The style matching loss is computed as shown in Formula 3:

$$L_{style} = \| G(\phi l(G(z, \varphi(t)))) - G(\phi l(x_{style})) \|_2^2 \quad (3)$$

Where $\phi l$ is the feature map from the l-th layer of a pre-trained convolutional network, $G(z, \varphi(t))$ is the generated image, and $x_{style}$ is the reference style image. $G(\cdot)$ represents the Gram matrix computation for the feature maps. By minimizing the difference between the Gram matrices of the generated image and the style image, the generator produces images with matching styles.Finally, the total loss function is the combination of the aforementioned losses as shown in Formula 4:

$$L_{total} = \lambda_{GAN}L_{GAN} + \lambda_{txt-img}L_{txt-img} + \lambda_{style}L_{style} \quad (4)$$

Where $\lambda_{GAN}$、$\lambda_{txt-img}$, and $\lambda_{style}$ are the weights that control the influence of each loss term on the final generated image.During training, we use an alternating optimization strategy to update both the generator and discriminator. In each iteration, we first fix the generator and update the discriminator's parameters to better differentiate between real and generated images. Then, we fix the discriminator and update the generator's parameters to produce more realistic images while ensuring semantic and style consistency.Through the design of these loss functions and optimization strategies, our model generates high-quality images that satisfy multimodal input requirements, ensuring consistency in both visual appearance and semantic understanding.

## 4 EXPERIMENTS AND RESULTS

To evaluate the effectiveness of our GAN-based multimodal image generation method, we conducted extensive experiments on several publicly available datasets. The experiments were designed to assess the visual quality, text-image consistency, and style matching performance of the generated images. Our evaluation focused on three primary dimensions: visual quality, text-image consistency, and style matching. The experimental process included dataset selection, metric design, experiment execution, and result analysis.We used two common multimodal image generation datasets: COCO Caption and Oxford-102 Flowers. The COCO Caption dataset includes complex real-world scenes with multiple text descriptions per image, making it suitable for evaluating the generator's performance in handling diverse scenes. The Oxford-102 Flowers dataset consists of detailed flower images with corresponding text descriptions, making it ideal for assessing the generator's ability to handle fine details and integrate style.In our experiments, the generator utilized a convolutional neural network (CNN) and a text encoder to produce high-resolution images. The discriminator evaluated the realism of the generated images and their semantic consistency with the text descriptions. We used the Adam optimizer with an initial learning rate of 0.0002 and a batch size of 64. The model was trained for 100 epochs on each dataset, with the Frechet Inception Distance (FID), Inception Score (IS), and text-image consistency scores calculated after each epoch.We evaluated the results in three aspects: visual quality, text-image consistency, and style matching performance.To assess the visual quality of the generated images, we used the Frechet Inception Distance (FID) and Inception Score (IS). FID measures the distribution difference between generated and real images (lower is better), while IS evaluates the clarity and diversity of generated images (higher is better). Table 1 shows the comparison of our method with existing methods on the COCO Caption and Oxford-102 Flowers datasets:

Table 1: Visual Quality Evaluation on COCO Caption and Oxford-102 Flowers (FID and IS)

| Dataset | Method | FID (↓) | IS (↑) |
| --- | --- | --- | --- |
| COCO Caption | Ours | 28.34 | 23.12 |
| COCO Caption | AttnGAN | 34.78 | 21.56 |
| COCO Caption | StackGAN | 38.12 | 20.83 |
| Oxford-102 Flowers | Ours | 12.76 | 4.92 |
| Oxford-102 Flowers | AttnGAN | 15.43 | 4.33 |
| Oxford-102 Flowers | StackGAN | 18.26 | 4.05 |

As shown in Table 1, our method achieves lower FID scores and higher IS scores across both datasets, indicating superior visual quality and diversity compared to AttnGAN and StackGAN. Notably, our model excels in generating images with fine details, especially in the Oxford-102 Flowers dataset, where the generated flowers closely resemble the real ones in terms of color, texture, and detail.To evaluate text-image consistency, we used the CLIP model to compute the semantic similarity between the generated images and text descriptions. CLIP maps both text and images to a shared feature space and calculates cosine similarity. Table 2 shows the text-image consistency scores on both datasets:

Table 2: Text-Image Consistency Evaluation on COCO Caption and Oxford-102 Flowers

| Dataset | Method | CLIP Consistency Score (↑) |
| --- | --- | --- |
| COCO Caption | Ours | 0.81 |
| COCO Caption | AttnGAN | 0.76 |
| COCO Caption | StackGAN | 0.74 |
| Oxford-102 Flowers | Ours | 0.89 |
| Oxford-102 Flowers | AttnGAN | 0.85 |
| Oxford-102 Flowers | StackGAN | 0.82 |

As shown in Table 2, our method achieves higher CLIP consistency scores on both datasets, demonstrating that the generated images align more closely with the semantic content of the text descriptions compared to existing methods.Style matching is one of the key features of our model. We designed a feature similarity experiment to quantify the alignment between the generated images and the reference styles. By comparing the high-level features of the generated images with those of the target style images, we computed the style matching scores. Table 3 presents the results:

Table 3: Style Matching Evaluation on COCO Caption and Oxford-102 Flowers

| Dataset | Method | Style Matching Score (↑) |
| --- | --- | --- |
| COCO Caption | Ours | 0.93 |
| COCO Caption | AttnGAN | 0.88 |
| COCO Caption | StackGAN | 0.85 |
| Oxford-102 Flowers | Ours | 0.95 |
| Oxford-102 Flowers | AttnGAN | 0.90 |
| Oxford-102 Flowers | StackGAN | 0.87 |

The results in Table 3 show that our model significantly outperforms others in style matching, with the generated images better reflecting the target style characteristics, especially in the Oxford-102 Flowers dataset. The generated flower images closely match the target styles in terms of color, texture, and overall aesthetics, validating the effectiveness of our style integration module.Our experiments demonstrate that the proposed method outperforms existing models in terms of visual quality, text-image consistency, and style matching. Our generator produces more realistic and semantically aligned images, handling complex multimodal inputs effectively. These results validate the effectiveness of our proposed architecture for multimodal image generation, showcasing its broad potential for practical applications.

## 5 CONCLUSION

This paper proposes a multimodal image generation method based on Generative Adversarial Networks (GAN), which effectively improves the quality, semantic consistency, and style diversity of generated images by integrating text, images, and styles. Experimental results show that our method outperforms existing approaches across multiple datasets, producing more realistic images that are semantically aligned with text descriptions and offering flexible style adjustments. This approach provides a novel solution for multimodal image generation and holds great potential for a wide range of applications.